\documentclass[runningheads]{llncs}
\usepackage[T1]{fontenc}
\usepackage{graphicx}

\begin{document}
\title{HeCiX: Integrating Knowledge Graphs and Large Language Models for Biomedical Research}
\titlerunning{HeCiX}
%
\newcommand{\equalcontrib}[1]{\textsuperscript{#1*}}
\newcommand{\authorfootnote}[1]{\let\thefootnote\relax\footnotetext{#1}}

\author{%
Prerana Sanjay Kulkarni\equalcontrib{1} \and 
Muskaan Jain\equalcontrib{2} \and 
Disha Sheshanarayana\equalcontrib{3} \and 
Srinivasan Parthiban\textsuperscript{4}
}
\institute{
  \inst{}Dept. of Computer Science and Engineering, PES University, Bengaluru, India\\
  \email{prer.kulk@gmail.com}
  \and
  \inst{}Dept. of Mathematics, IIT Madras, Chennai, India \\
  \email{muskaan40389@gmail.com}
  \and
  \inst{}Dept. of Computer Science and Engineering, Manipal University, Jaipur, India\\
  \email{disha.229301161@muj.manipal.edu}
  \and
  \inst{}Dept. of Data Science and Engineering, Indian Institute of Science Education and Research, Bhopal, India\\
  \email{parthi@iiserb.ac.in}
}
\maketitle  
\authorfootnote{* These authors contributed equally to this work.}
\begin{abstract}
Despite advancements in drug development strategies, 90\% of clinical trials fail. This suggests overlooked aspects in target validation and drug optimization. In order to address this, we introduce HeCiX-KG, Hetionet-Clinicaltrials neXus Knowledge Graph, a novel fusion of data from ClinicalTrials.gov and Hetionet in a single knowledge graph. HeCiX-KG combines data on previously conducted clinical trials from ClinicalTrials.gov, and domain expertise on diseases and genes from Hetionet. This offers a thorough resource for clinical researchers. Further, we introduce HeCiX, a system that uses LangChain to integrate HeCiX-KG with GPT-4, and increase its usability. 
HeCiX shows high performance during evaluation against a range of clinically relevant issues, proving this model to be promising for enhancing the effectiveness of clinical research. Thus, this approach provides a more holistic view of clinical trials and existing biological data.

\end{abstract}
\section{Introduction}
\label{sec:introduction}
The pharmaceutical industry faces significant challenges in drug discovery, with alarming clinical trial failure rates of almost 90\% ~\cite{fails}. The rise in attrition rates reflects not only huge financial losses but also delayed implementation of life-saving treatments for patients.
 
One of the major reasons underlying this is the fragmented nature of the available data. Hetionet~\cite{himmelstein2017} contains vast domain knowledge about diseases, genes, and anatomy, yet it lacks sufficient information about previously conducted clinical trials and experiments. Conversely, ClinicalTrials.gov~\cite{ctgov} houses extensive information about clinical trials and experiments conducted worldwide, including details about Principal Investigators of studies which can be useful in identifying Key Opinion Leaders (KOLs). However, it offers limited insights into the diseases themselves. This disparity between our understanding of fundamental biology and clinical trial outcomes hinders effective drug development.

To address this, we propose a novel knowledge graph, HeCiX-KG (Hetionet-Clinicaltrials neXus Knowledge Graph), which integrates information from clinicaltrials.gov~\cite{ctgov} and Hetionet~\cite{himmelstein2017}. HeCiX-KG is a single knowledge graph, connecting biological knowledge with clinical trial data. This integration can provide better understanding, revealing linkages and patterns that were previously missed, but are vital for the effective repurposing and discovery of new drugs.

Building upon HeCix-KG, we introduce HeCiX, a system that utilizes OpenAI's GPT-4~\cite{gpt4} and LangChain~\cite{langchain} to enable seamless interaction with the knowledge graph. HeCiX translates natural language queries into CQL (Cypher Query Language) queries, which makes it possible to efficiently retrieve relevant context from the knowledge graph. Subsequently, the system displays the result in human-understandable format, rendering the information available to clinical and biomedical researchers.

We evaluate HeCiX's performance against a wide array of question-answering tasks relevant to the domain. The results show notable advancements in the scope and depth of data obtained, thus providing a helpful tool to enhance the efficiency of clinical research, and improve the success rates of drug repurposing and development. HeCiX overcomes significant shortcomings in the current resources by offering a holistic view of disease biology, clinical trial history, and expert knowledge at the user's fingertips.

The structure of the paper is as follows. We describe the background of the work and the knowledge graph construction in Section 2 and Section 3 respectively. Section 4 talks about the detailed methodology. Experimentation is described in Section 5 which is followed by results and discussions in Section 6. Finally, Section 7 discusses the conclusion.

\section{Background and Related Work}
\label{sec:background}

\subsection{Knowledge Graphs in Biomedical Research}
Knowledge graphs have emerged as powerful tools for representing and integrating complex biomedical information, facilitating efficient data integration and knowledge discovery. Notable examples include Bio2RDF~\cite{bio2rdf}, CTKG by Chen et al.~\cite{ctkg}, Hetionet~\cite{himmelstein2017}, and others. They have been used to enhance drug discovery, understand disease mechanisms, and identify biomedical relationships.

\subsection{Large Language Models in Healthcare}
Large Language Models (LLMs) have significantly improved medical literature analysis, clinical note interpretation, and diagnosis support. Important models in this field include DeepMind's MedIC, Microsoft's BioGPT~\cite{biogpt}, TrialGPT~\cite{trialgpt}, BioBart~\cite{biobart}, and BioMistral~\cite{biomistral}, among others. They have impacted domains such as drug discovery, patient data analysis, and clinical decision support.

\subsection{Hetionet: A Comprehensive Biomedical Knowledge Graph}
Hetionet is a heterogeneous network of biomedical knowledge, integrating data such as genes, compounds, diseases, and their interrelationships~\cite{himmelstein2017}. It has been used in predicting drug-target interactions, identifying disease mechanisms, and supporting drug repurposing efforts, making it ideal for integration with other data sources to enhance clinical research capabilities.

\subsection{ClinicalTrials.gov: A Repository of Clinical Trial Data}
ClinicalTrials.gov is an extensive data source providing information about clinical trials, studies conducted on various diseases, principal investigators, study locations, and trial outcomes~\cite{ctgov}. It thereby supports clinical research and drug development.

\section{Knowledge Graph Construction}
\label{sec:kgconstruction}

HeCiX-KG is constructed from two primary sources of data, Hetionet~\cite{himmelstein2017} and ClinicalTrials.gov~\cite{ctgov}. It combines their data into a single knowledge source and includes data related to six specific diseases, namely Vitiligo, Atopic Dermatitis, Alopecia Areata, melanoma, Epilepsy, and Hypothyroidism.

\subsection{Hetionet}

Hetionet is a highly interconnected knowledge base, which combines data from 29 distinct databases. It comprises a total of 47,031 nodes across 11 types: Disease, Compound, Gene, Symptom, Side Effect, Biological Process, Molecular Function, Anatomy, Cellular Component, Pathway, and Pharmacologic Class~\cite{himmelstein2017}. 
For the purpose of constructing HeCiX-KG, we have extracted a subgraph of Hetionet, consisting of data related to the six chosen diseases. This comprises a total of 1071 nodes and 1125 relations.

\subsection{ClinicalTrials.gov}

ClinicalTrials.gov provides massive amounts of information about clinical trials and studies on various diseases and conditions~\cite{ctgov}. While the total number of records in ClinicalTrials.gov exceeds 500,000, our research focuses on a selected subset of 1,200 records, spanning the six selected diseases. This subset when constructed as a knowledge graph, contains 5,454 nodes and 11,466 edges. The nodes in this subset are classified into 9 types: Disease, Principal Investigators (PI), Study, Conditions, Phases, Locations, Interventions, Age, and Sex. There are 10 types of relationships connecting these nodes.

\subsection{Schema}

\begin{figure}[h]
    \centering
    \includegraphics[width=\textwidth]{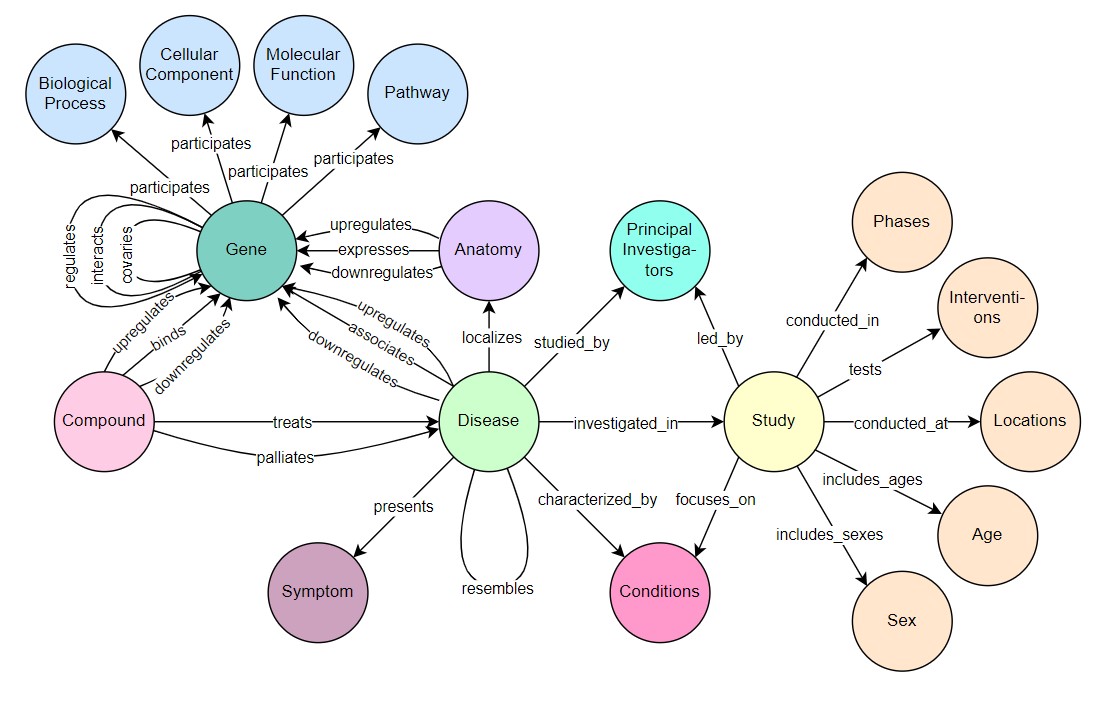}
    \caption{HECIX-Schema}
    \label{fig:hecix-schema}
\end{figure}

 By taking inspiration from the schemas of the knowledge graphs constructed in Hetionet~\cite{himmelstein2017} and the work of Devarakonda et al.~\cite{novartis-kg}, we have constructed a comprehensive schema for HeCiX-KG. The `Disease' node serves as the main connecting point between the two integrated databases. This schema has been illustrated in Figure \ref{fig:hecix-schema}. By populating the schema with our data, we have obtained a knowledge graph consisting of 6,509 nodes and 14,377 edges.

\section{Methodology}
\label{sec:methodology}

Our approach consists of two major stages, construction of HeCiX-KG, and its integration with GPT-4 using LangChain.

\subsection{Knowledge Graph Construction}
HeCiX-KG is constructed by extracting and integrating relevant data from Hetionet~\cite{himmelstein2017} and ClinicalTrials.gov~\cite{ctgov} for six specific diseases. The resulting knowledge graph has 6,509 nodes and 14,377 edges. The construction process involves data extraction, schema design, entity-relationship mapping, and graph population. 

\subsection{LLM Integration using LangChain}

\begin{figure}[htbp]
    \centering
    \includegraphics[width=1\textwidth]{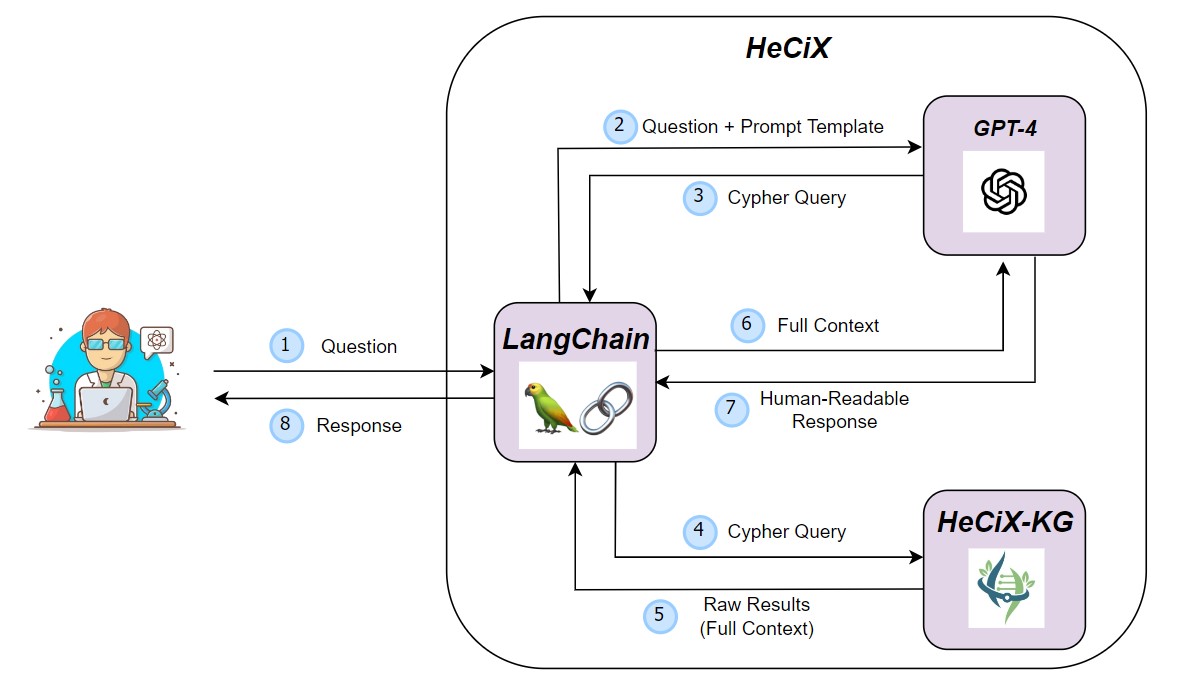}
    \caption{Query processing pipeline}
    \label{fig:Query processing pipeline}
\end{figure}

To enhance the usability of HeCiX-KG, we developed HeCiX, a system that integrates our knowledge graph with GPT-4 using LangChain. Specifically, we utilized the GraphCypherQAChain component from the LangChain ecosystem for this integration. As indicated in Figure \ref{fig:Query processing pipeline}, our query processing pipeline is as follows:

\begin{enumerate}
    \item \textbf{User Query Input}: A user submits a natural language prompt to LangChain.
    
    \item \textbf{Query and Prompt Processing}: The user's question is combined with a set prompt template, and then sent to GPT-4
    
    \item \textbf{Cypher Query Generation}: GPT-4 generates a Cypher query based on the user's input and sends it back to LangChain.
    
    \item \textbf{Database Querying}: LangChain executes the generated Cypher query on HeCiX-KG.
    
    \item \textbf{Raw Results Retrieval}: HeCiX-KG returns the raw query results (the `Full Context') to LangChain.
    
    \item \textbf{Context Forwarding}: LangChain forwards the full context to GPT-4 for interpretation and conversion into a human-readable format.
    
    \item \textbf{Human-Readable Response Generation}: GPT-4 generates a human-readable response based on the full context sent to it, and sends it to LangChain.
    
    \item \textbf{User Response}: Finally, LangChain returns the human-readable response to the user, thereby providing the user with the answer to their query.
\end{enumerate}

This integration allows users to interact with the complex HeCiX-KG using natural language queries, significantly enhancing its accessibility and usability for clinical researchers.

\section{Experimentation}
\label{sec:experimentation}

\subsection{Experimental Setup}

Our experimental setup consisted of the following major steps:

\begin{enumerate}
    \item We set up our AuraDB~\cite{neo4j_auradb} server to host HeCiX-KG.
    \item We constructed the schema of our knowledge graph, based on the individual structures of Hetionet and clinical trials data.
    \item We populated the schema with data from both Hetionet and ClinicalTrials.gov to create HeCiX-KG.
    \item We integrated HeCiX-KG with GPT-4 using LangChain's GraphQueryQAChain component.
\end{enumerate}

\subsection{Evaluation Methodology}

To properly assess and analyse the question-answering abilities of our system, we designed a set of question-answering tasks for HeCiX to answer. These tasks were carefully crafted to cover various aspects of clinical research, including drug discovery, identification of KOLs, and analysis of biomedical data, among others.

\section{Results and Discussion}
\label{sec:resultsanddiscussion}

\subsection{Experimentation Results}

We utilized the RAGAS~\cite{ragas} framework to evaluate our model's performance. The framework calculates several key metrics: faithfulness, answer relevance, context precision, and context recall. The results of our experimentation have been displayed in Table 1.

\begin{table}[h]
\centering
\caption{Performance metrics of HeCiX using the RAGAS framework}
\begin{tabular}{|l@{\hspace{0.5em}}|@{\hspace{0.5em}}c@{\hspace{0.5em}}|}
\hline
\rule{0pt}{2.5ex} \textbf{Metric} & \textbf{Score} \\[0.5ex]
\hline
\rule{0pt}{2.2ex} Faithfulness & 0.8572 \\[0.5ex]
\rule{0pt}{2.2ex} Answer Relevance & 0.9340 \\[0.5ex]
\rule{0pt}{2.2ex} Context Precision & 0.9202 \\[0.5ex]
\rule{0pt}{2.2ex} Context Recall & 0.6654 \\[0.5ex]
\hline
\end{tabular}
\label{tab:performance_metrics}
\end{table}

\subsection{Comparative Analysis}

In addition to evaluating our model on the evaluation metrics seen earlier, we also carried out a comparative analysis with OpenAI's GPT-4 and Anthropic's Claude 3 Sonnet~\cite{anthropic_claude_sonnet} in a zero-shot scenario. We tested these models on the same set of questions used to evaluate HeCiX. We discovered that while these models excel at answering questions based on domain-specific knowledge, they struggle with queries related to clinical trials.

\begin{figure}[htbp]
    \centering
    \includegraphics[width=1\textwidth]{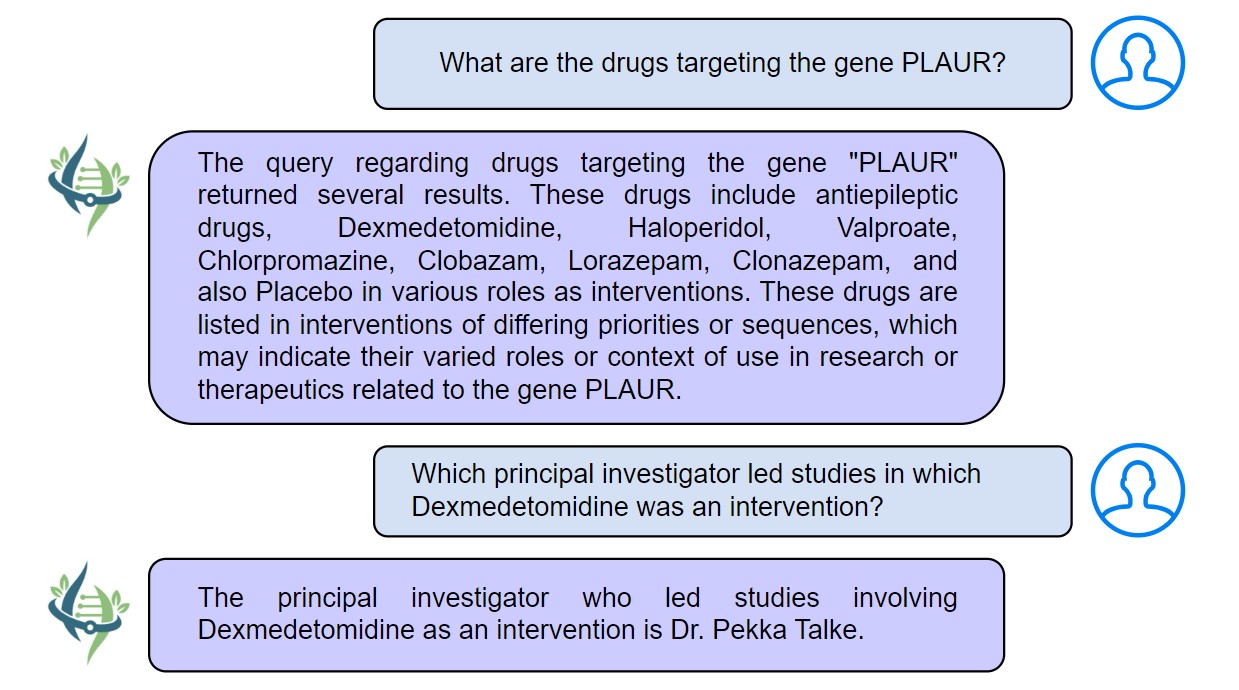}
    \caption{Sample interaction with HeCiX demonstrating QA capabilities}
    \label{fig:results}
\end{figure}

Our experiments with HeCiX demonstrated promising results in enhancing clinical research capabilities. Figure \ref{fig:results} shows a sample interaction with HeCiX, where HeCiX is answering questions which test both domain expertise and knowledge about clinical trials. 

\subsection{Limitations}
While HeCiX has shown promising results, it is important to acknowledge its limitations.

\begin{itemize}
    \item Uncertainty of the model's performance as the knowledge graph expands.
    \item Additional testing on a wider range of diseases to ensure robustness of the system.
\end{itemize}






\section{Conclusion}
\label{sec:conclusion}

This paper introduces HeCiX, an innovative system that connects knowledge graphs from clinical trials data and Hetionet with large language models to address major challenges in clinical research. Our experiments showcase that HeCiX enhances drug discovery processes and uses existing scattered biomedical data effectively. HeCiX uncovers all possible relationships between diseases, genes, and treatments, potentially accelerating drug development which can lead to unexpected discoveries.

HeCiX is a major advancement in using AI for clinical research. We believe that HeCiX will play a crucial role in shaping the future of biomedical research, providing a connected and innovative ecosystem in the field.

The scope of future enhancements for HeCiX could include adding SNOMED CT to the knowledge graph for better clinical term standardization and employing meta-path matching techniques to identify complex relationships more effectively.

%
%
%
%

\end{document}